\documentclass[preprint,5p, twocolumn]{elsarticle}

\usepackage{amssymb}
\usepackage{amsmath}
\usepackage{booktabs}
\usepackage{acronym}
\usepackage{subfigure}
\usepackage{subcaption}
\usepackage{color,soul}
\acrodef{GQN}{Generative Query Network}
\acrodef{MAE}{Mean Absolute Error}
\acrodef{MAPE}{Mean Absolute Percentage Error}
\acrodef{MVCNN}{Multi-view Convolutional Neural Network}
\acrodef{SMAPE}{Symmetric Mean Absolute Percentage Error}
\journal{Manufacturing Letters}

\begin{document}

\begin{frontmatter}



\title{Technology prediction of a 3D model using Neural Network}


\affiliation[m]{organization={Masta Solutions sp. z o.o.},%Department and Organization
            city={Poznan},
            country={Poland}}

\affiliation[p]{organization={Institute of Computing Science, Poznan University of Technology},%Department and Organization
            city={Poznan},
            country={Poland}}

\author[m,p]{Grzegorz Miebs\corref{cor}}
\cortext[cor]{Corresponding author: email gmiebs@cs.put.poznan.pl}

\author[p]{Rafał A. Bachorz}

\begin{abstract}
Accurate estimation of production times is critical for effective manufacturing scheduling, yet traditional methods relying on expert analysis or historical data often fall short in dynamic or customized production environments. This paper introduces a data-driven approach that predicts manufacturing steps and their durations directly from 3D models of products with exposed geometries. By rendering the model into multiple 2D images and leveraging a neural network inspired by the Generative Query Network, the method learns to map geometric features into time estimates for predefined production steps with a mean absolute error below 3 seconds making planning across varied product types easier.
\end{abstract}


% \begin{highlights}
% \end{highlights}

\begin{keyword}
Technology Prediction \sep Neural Network \sep Machine Learning \sep Deep Learning

\end{keyword}

\end{frontmatter}


\section{Introduction}
\label{sec:sample1}
Accurate production scheduling is a cornerstone of efficient manufacturing. In practice, schedules are generated based on estimates of processing times required for each step in the production process. However, when these estimates deviate from actual conditions—due to missing or outdated data - the generated schedules quickly become obsolete. This mismatch leads to inefficiencies, frequent rescheduling, and, ultimately, a decline in overall manufacturing performance.\citep{liMachineLearningOptimization2020a}

Traditionally, estimating production times has required manual analysis by experts or historical data from similar products. While effective in some cases, these methods are time-consuming, error-prone, and difficult to scale, particularly in environments where products are frequently customized or modified. For instance, Żywicki and Osiński discuss the challenges manufacturers face in calculating production times for customized products, highlighting the lack of real data on standard times required for individual manufacturing operations. \citep{10.1007/978-3-030-18789-7_11, 10.1007/978-3-319-91334-6_51}

This research proposes an automated approach to predicting production steps such as welding, cutting, bending, etc. and their corresponding times based solely on the 3D model of a product. The central idea is to train a neural network that learns to map geometric features of a product - extracted from rendered images of its 3D model - into a vector of processing times across a predefined set of manufacturing steps.

The proposed method adapts the Generative Query Network (GQN)\citep{Eslami2018, Miebs2022}, representing 3D models as 
collections of 2D images taken from various viewpoints 
% RAB
called a scene. 
%RAB
These images serve as input to an encoder-decoder neural network architecture: the encoder, based on EfficientNetB0 architecture \citep{pmlr-v97-tan19a}, extracts meaningful latent representations from individual images, and the decoder aggregates and transforms these into time estimates for each production step. This architecture is flexible, allowing the use of a variable number of input images, and scalable to different product types without manual feature engineering.

Using historical manufacturing data as an input, this model offers a promising direction toward data-driven process planning, enabling faster, more adaptive, and more precise scheduling in modern production systems. 

The key contribution of this work is the adaptation of the GQN concept~\citep{Eslami2018} to the problem of predicting production times. We successfully designed and trained a neural model that leverages the EfficientNet architecture~\citep{pmlr-v97-tan19a}. This proved to be a reliable approach, outperforming several other selected methods.

\section{Methods}
\label{sec:sample:appendix}
\subsection{Task}
The aim is to predict a vector $t_i$ of length $k$ denoting the times required at each step $j,j\in\{0,1,...,k-1\}$ to produce the item $i$ based on its 3D model ($m_i$). The $k$ steps are predefined and constant for each $i$. The items can be divided into a reference set $A^R$ for which both $m$ and $t$ are available and a non-reference set $A^{NR}$ for which only the 3D models are available but times are unknown. The reference set consists of items already produced or analyzed by experts, hence the production times and steps are known. Formally, the task is to create a function $f$ that maps $m$ into $t$, $f(m) = t$. 

\subsection{Data Processing}
\begin{figure*}[htbp]
    \centering

    % Top subfigure with caption and label
    \subfigure[Procedure of images generation.]{%
        \includegraphics[width=0.55\textwidth]{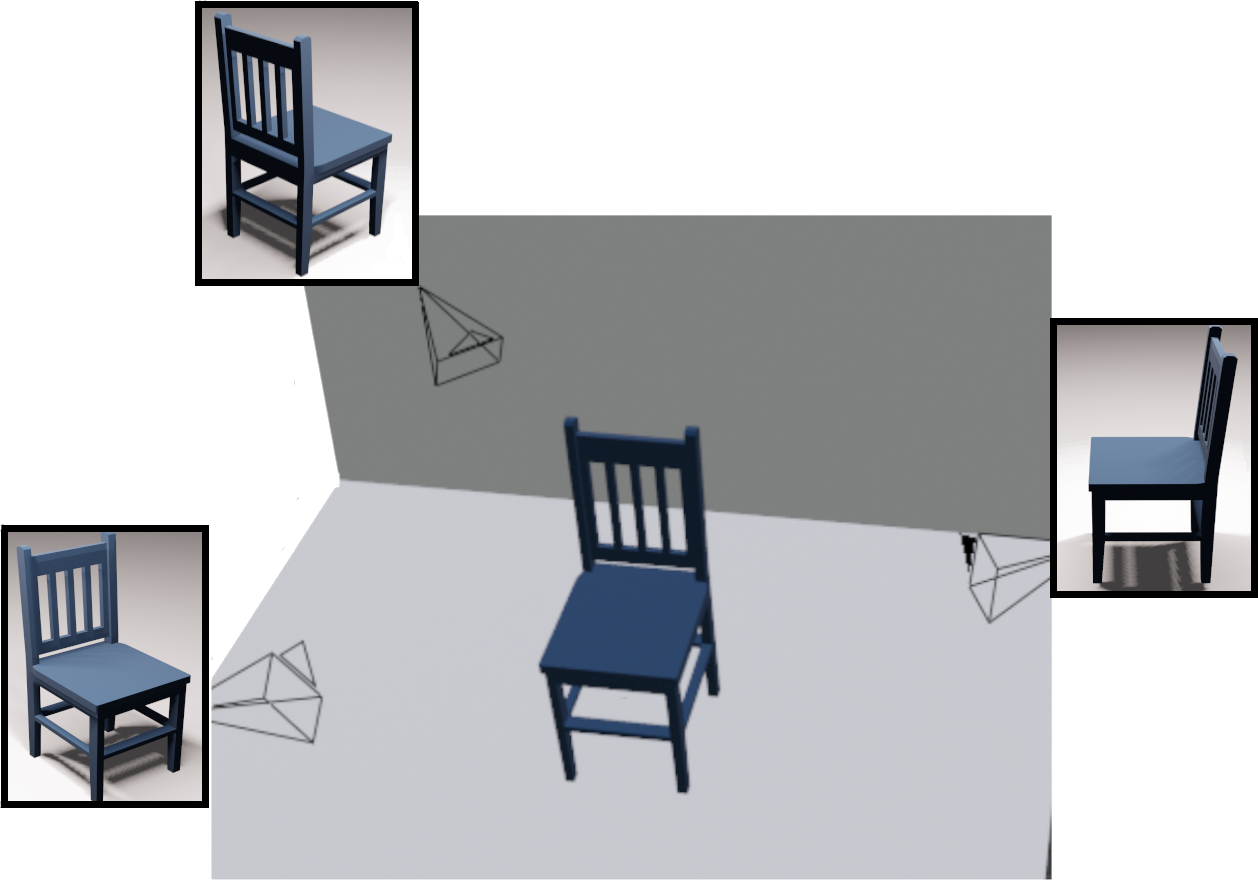}
        \label{fig:images}
    }

    \vspace{1em} % Optional vertical spacing

    % Bottom subfigure with caption and label
    \subfigure[High-level perspecticve of proposed neural model.]{%
        \includegraphics[width=0.75\textwidth]{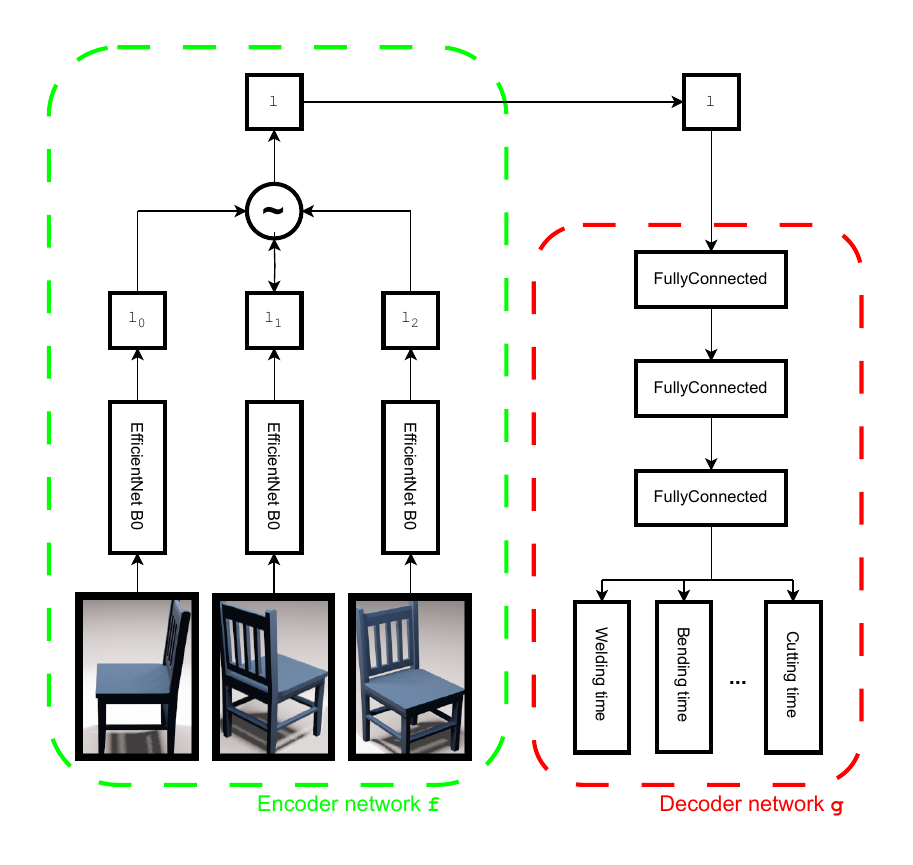}
        \label{fig:neural}
    }
    % Reviewer #1, remark 2.
    \caption{The prediction pipeline reflecting a high-level idea behind Generative Query Network. The upper panel (a) illustrated the idea of scene creation, i.e. reflecting the 3D object as a collection of 2D images. The bottom part introduces the adaptation of GQN architecture more formally. The network is subdivided into the Encoder and Decoder networks, marked green and red, respectively. The former turns the arbitrary number of images into the fixed-length latent space, whereas the Decoder network, on the other hand, accepts the latent space representation, applies further fully-connected transforms, and predicts the time estimates for each production step.}
    \label{fig:main-figure}
\end{figure*}

The proposed architecture is inspired by the \ac{GQN} \citep{Eslami2018} approach. In that model, a 3D scene is represented as a set of 2D images picturing the scene from different angles. That procedure was adopted for the technology prediction task. The process is presented in Figure~\ref{fig:images}. To represent an item to be produced, a set of images $I = \{i_0, i_1, \ldots \}$ is generated. They are taken from different camera positions $P = \{p_0, p_1, \ldots \}$ to capture each detail of the item. This is possible because all components of the considered objects are visible from the outside. The camera can take any position on a spherical shell surrounding the object, as long as it is oriented toward the object's center, the entire object is visible, and its minimal bounding box covers at least 50\% of the image. In the presented approach the number of images was randomly selected between $3$ and $5$. To describe the position of a camera $3$ numbers were used denoting its position alongside $x$, $y$, and $z$ axes and additional $3$ numbers informing about its rotation on these axes.

\subsection{Neural architecture}
The neural architecture of the proposed model, presented in Figure~\ref{fig:neural}, is based on the encoder-decoder approach. The former is responsible for transforming an image into a fixed-size latent vector representing relevant information from the image~\citep{Ji2021}. The latter is used to change the latent vectors into meaningful predictions, in this case, production times. In the proposed model an EfficientNetB0~\citep{pmlr-v97-tan19a}, a state-of-the-art machine learning model for image processing tasks, is used as the encoder while a simple multilayer perceptron network with 3 layers, consisting of 256, 128, and 32 neurons, serves as the decoder. The encoder $f^\prime$ is applied independently to each input image, producing a latent vector for each of them $L = \{l_0, l_1, ... \}$, $f^\prime(i_i) = l_i$. Those vectors are then aggregated using a piecewise mean operator to create a vector $l$ containing information regarding the whole 3D model. This procedure allows to use any number of pictures within the same network since the architecture and the number of trainable parameters do not depend on the number of input images. Vector $l$ is then processed by the decoder $g$ consisting of simple classic fully connected layers. The output layer consists of $k$ neurons with a ReLU activation function returning a predicted vector representing the time required at each production step, where a value of $0$ indicates that the step is not required.

\section{Experiment}
The dataset in the experiment consists of 539 3D models of items with corresponding vectors denoting times of $k=6$ different manufacturing stages (welding, cutting, bending, screwing, drilling, and assembling). If production of a given item does not require all steps then for the unnecessary stages the time is set to zero. Sample times are presented in Table~\ref{tab:example1} and selected objects are presented on Figure~\ref{fig:snaps}. Randomly chosen $60$ of these objects were put aside as a test set used to evaluate the model, while the remaining $479$ items were used for training purposes. For each training object a hundred snapshots were generated. During the training in each epoch a different subset was selected to represent an item. This procedure was essential due to limited number of training examples. In this data augmentation process number of unique examples coming from one item was close to 80 million ${100\choose 3} + {100\choose 4} + {100\choose 5} = 79370445$. However, they cannot be treated as independent training examples since they all originate from the same 3D object, nevertheless, such procedure allows preventing overfitting of the model, thus attaining better results. This augmentation method was compared with a classical image augmentation technique where only one set of views is generated per object. To create more training samples images are shifted, rotated, and zoomed. This approach is later referred to as GQN-static.

The neural network was implemented using TensorFlow library and for rendering Blender software was used \citep{tensorflow2015-whitepaper, blender}. Training the model took 8 hours using two NVIDIA Titan RTX GPUs.

\begin{table}
\footnotesize 
    \centering
    \caption{Example matrix presenting manufacturing times in seconds of selected items}
    \label{tab:example1}
    \addtolength{\tabcolsep}{-0.2em}
    \begin{tabular}{lcccccc}
\toprule
ID & welding & cutting & bending & screwing & drilling & assembling \\
\midrule
1 & 0 & 0 & 20 & 0 & 200 & 0 \\
2 & 300 & 60 & 0 & 0 & 0 & 0 \\
3 & 0 & 30 & 5 & 0 & 15 & 0 \\
4 & 0 & 60 & 0 & 0 & 0 & 60 \\
5 & 0 & 0 & 10 & 0 & 260 & 0 \\
6 & 60 & 30 & 0 & 0 & 30 & 0 \\
7 & 0 & 30 & 0 & 60 & 0 & 0 \\
8 & 300 & 60 & 0 & 0 & 0 & 0 \\
9 & 0 & 60 & 0 & 0 & 0 & 300 \\
\bottomrule
\end{tabular}
\end{table}

\begin{figure*}[h]
    \centering
    \setlength{\tabcolsep}{1pt} % Reduce space between columns
    \renewcommand{\arraystretch}{0} % Reduce space between rows
    \begin{tabular}{ccc}
    1 & 2 & 3 \\
        \subfigure{\includegraphics[width=0.25\textwidth]{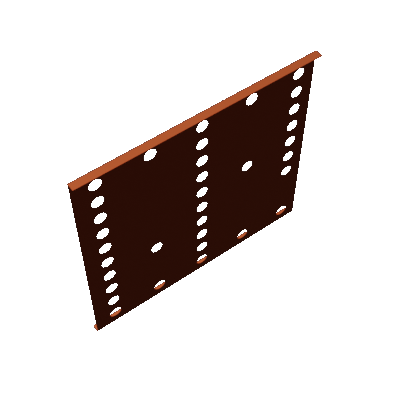}} &
        \subfigure{\includegraphics[width=0.25\textwidth]{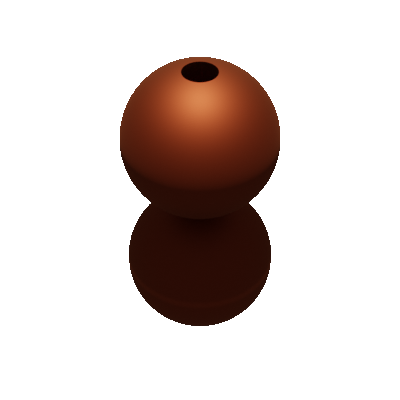}} &
        \subfigure{\includegraphics[width=0.25\textwidth]{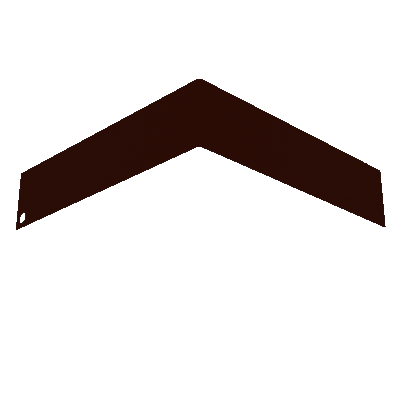}} \\
        
        \subfigure{\includegraphics[width=0.25\textwidth]{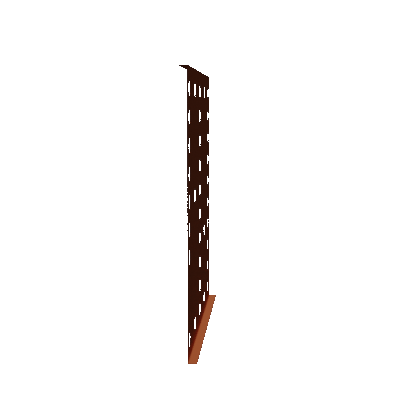}} &
        \subfigure{\includegraphics[width=0.25\textwidth]{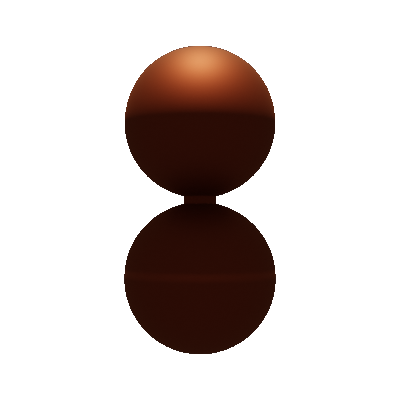}} &
        \subfigure{\includegraphics[width=0.25\textwidth]{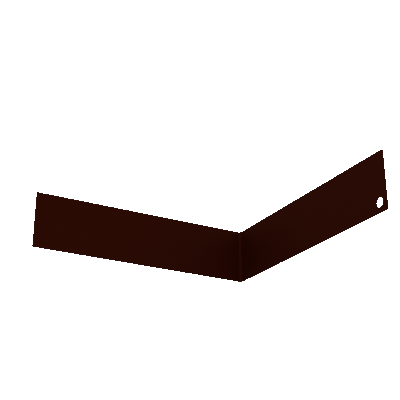}} \\

        \subfigure{\includegraphics[width=0.25\textwidth]{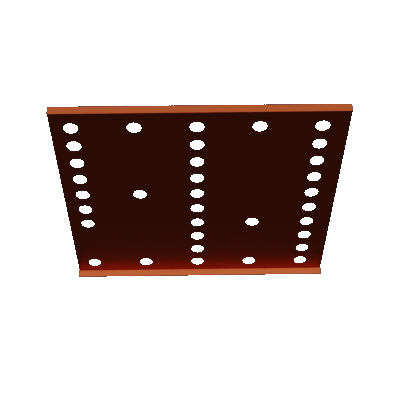}} &
        \subfigure{\includegraphics[width=0.25\textwidth]{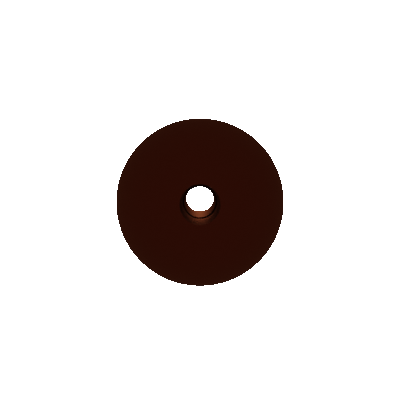}} &
        \subfigure{\includegraphics[width=0.25\textwidth]{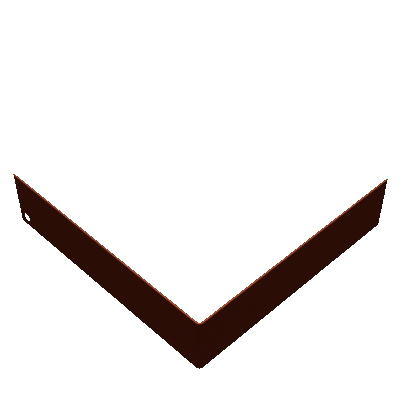}} \\
    \end{tabular}
    \caption{Example snapshots of objects used to train the model.}
    \label{fig:snaps}
\end{figure*}

\begin{table}
    \centering
    \caption{Performance comparison including mean absolute error together with a standard deviation, mean absolute percentage error, f1 score for binary classification, and symmetric mean absolute percentage error}
    \begin{tabular}{@{}lcccc@{}}
        \toprule
        Model   & MAE   & MAPE & F1 & SMAPE   \\ \midrule
        GQN & $2.76 \pm 12.34$  & 7.8 & .93 & 6.72  \\
        PointNet & $16 \pm 31.17$  & 18.66 & .87 & 13.48 \\ 
        MVCNN & $2.8 \pm 12.3$  & 8.21 & .93  & 7.06 \\ 
        GQN-static & $3.05 \pm 13.77$ & 8.51 & .88 & 12.33  \\\bottomrule
    \end{tabular}
    \label{tab:model_performance}
\end{table}

To verify the quality of the proposed approach a PointNet~\citep{https://doi.org/10.48550/arxiv.1612.00593} model was trained on the point clouds generated from the 3D models of the same dataset. A similar approach was successfully applied to such problems as detecting geometric defects in gears~\citep{Mei2024}, identification of mechanical parts for robotic disassembly~\citep{Zheng2022}, or classification of crystalline structures in complex plasmas~\citep{Dormagen2024}.  For this model and data type, there was no straightforward way to augment the data, hence the PointNet model training was significantly faster, but the model was not able to learn how to map a point cloud to production times. The other neural architecture used for reference was a \ac{MVCNN}~\citep{mvcnn}. The approach is similar to the proposed GQN, however, it uses a convolutional network also in the decoder phase and does not use the position of cameras. Results obtained by the models are presented in Table~\ref{tab:model_performance}. The \ac{MAE} for results obtained using the \ac{GQN} approach is equal to 2.76 seconds. The \ac{MAPE} of this model is equal to $7.8\%$. To calculate \ac{MAPE} when the expected value is zero the following procedure was applied
\[
mape(real, pred)= 
\begin{cases}
    0\%, & \text{if } real = 0 \land pred < 1 \\
    100\%, & \text{if } real = 0 \land pred \geq 1 \\
    \frac{|real - pred|}{real}, & \text{otherwise}
\end{cases}
\]
meaning predicting more than 1 second if the real value is zero results in 100\% error while predicting less than a second if a step is not needed results in 0\% error. This approach is motivated by a reasonable assumption that no manufacturing process can be done in less than a second. The model achieved \ac{SMAPE} defined as $\frac{100}{n}\sum\frac{|real - pred|}{|rea| + |pred|}$ of $6.72\%$. Finally, to evaluate capabilities to predict whether a step is needed or not, the F1 score was applied. Predicted production times were binarized with a cut-off set to 1 second. The proposed model achieved F1 score of $0.93$.  
The PointNet model was over 5 and 2 times worse in terms of, respectively, MAE and MAPE, compared to models utilizing 2D images. The GQN model performed comparably with MVCNN on the MAE metric. The former is slightly better at predicting whether a production step is needed which is reflected in MAPE and SMAPE metrics. A model with classical image segmentation performed worse than models using different snapshots.

\section{Conclusions}
The proposed neural network is capable of learning the necessary steps and their durations to produce an item with exposed geometries, and it outperforms the PointNet model. The excellent results were facilitated in part by the simplicity of the considered items, which featured basic shapes with no hidden details. In the case of more complex objects, representing an item using snapshots will not allow for capturing all the necessary information. As a result, the GQN model will fail to accurately predict the required tasks and their durations. The second crucial aspect was the limited number of training examples. Regarding this aspect, the proposed approach with a straightforward data augmentation strategy had a huge advantage. 
Further study with significantly more 3D models including those more complex and with hidden details is planned. In that scenario, more complex approaches will be considered e.g. extending the current approach by using not only snapshots but also cross sections and comparing it with a PointNet++~\citep{https://doi.org/10.48550/arxiv.1706.02413} model.

\section*{CRediT authorship contribution statement}
\textbf{Grzegorz Miebs:} Writing - Original Draft, Investigation, Methodology, Software. \textbf{Rafa\l~ A. Bachorz:} Writing - Review~\&~Editing, Investigation, Methodology, Software.

\section*{Declaration of competing interest}
The authors declare that they have no known competing financial interests or personal relationships that could have appeared
to influence the work reported in this paper.

\section*{Acknowledgements}
We thank Patrycja Nowak for designing the graphics used in the paper.
This project has received funding from the European Union’s Horizon 2020 research and innovation programme under grant agreement N. 101017057

 % \bibliographystyle{elsarticle-num} 
 % \bibliography{cas-refs}


\end{document}